\documentclass[11pt]{article}
\usepackage{geometry}
\usepackage{coling2020}
\usepackage{times}
\usepackage{url}
\usepackage{latexsym}
\usepackage{microtype}
\usepackage{footmisc}
\usepackage{caption}
\usepackage{graphicx, subfig}
\usepackage{amsfonts}

\colingfinalcopy 

\title{KaLM at SemEval-2020 Task 4: Knowledge-aware Language Models for Comprehension And Generation}
\author{Jiajing Wan* \quad Xinting Huang*  \\
	\\
	Sun Yat-Sen University \\
	Guangzhou, China\\
	{\tt \{wanjj, huangxt39\}@mail2.sysu.edu.cn} \\}

\date{}

\begin{document}
	\maketitle
	\begin{abstract}
			\blfootnote{
			*denotes equal contribution.
			
			%
			%
			%
			%
			%
			
			\hspace{-0.65cm}  
			This work is licensed under a Creative Commons 
			Attribution 4.0 International License.
			License details:
			\url{http://creativecommons.org/licenses/by/4.0/}.
		}
	
		This paper presents our strategies in SemEval 2020 Task 4: Commonsense Validation and Explanation. We propose a novel way to search for evidence and choose the different large-scale pre-trained models as the backbone for three subtasks. The results show that our evidence-searching approach improves model performance on commonsense explanation task. Our team ranks 2nd in subtask C according to human evaluation score.\footnote{The code is available at https://github.com/huangxt39/KaLM}
	\end{abstract}
	
	\section{Introduction}
	
	Commonsense reasoning has been seen as one of the key ability for intelligent machines to perform various activities \cite{3ddd816b11cd42a7a49f624dc0206a8e}. SemEval 2020 Task 4 \cite{wang-etal-2020-semeval} is a commonsense validation and explanation task which is inspired by Wang et al. \shortcite{DBLP:journals/corr/abs-1906-00363}. This task consists of three subtasks. The first subtask is to choose the one which makes sense from two natural language statements with similar wordings; The second subtask is to decide among three options the most crucial reason why a given statement does not make sense; The third subtask requires the machine to generate the reasons. 
	
	
	To make predictions or generate reasons, background knowledge is essential. A simple way to supplement that knowledge is utilizing plain texts from natural language databases, e.g. Wikipedia. Intuitively, for a specific given statement, plain texts can be provided by searching similar sentences in the database. In other words, each evidence sentences contain the keywords of the given statement. This method is used in some of the state-of-the-art models \cite{lv2019graphbased}. However, for the purpose of explaining the reason, evidence which has similar wording with the given statement may lack information, and sometimes can be misleading when the given statement does not make sense.
	
	In this work, we propose a novel way for evidence-searching using plain texts. We obtain evidence by searching for the meaning of the keywords in the given statement. In other words, using evidence containing the meanings of the keywords rather than containing the keywords themselves. The reason for our method is that these meanings may provide important information to explain why a statement makes sense or does not make sense. For example in Figure~\ref{image:evidence}, the definition of "aircraft carrier"---"A warship designed to carry aircraft"---is given by the evidence obtained from the database. Such a definition well explains why the given statement is wrong and can be used when generating the reason. In contrast, evidence containing both "aircraft carrier" and "human" will not be helpful.
	
	We conduct experiments on subtask A, B, and C. Results show that our evidence-searching method boosts the performance on subtask C. Our team achieves accuracy of 95.3 (9th place) in subtask A and 93.2 (7th place) in subtask B. In subtask C, Our approach achieves the BLEU score of 18.5 (3rd place) and human evaluation score of 2.08 (2nd place). Moreover, the best BLEU score in our experiments (20.4) even outperforms the score we obtain in competition.
	
	\begin{table}[]
		\begin{tabular}{lll}
			\hline
			Task      & Description                                                                                                                 & Question example                                                                                                                                                                                                                                 \\ \hline
			Subtask A & \begin{tabular}[c]{@{}l@{}}Choose the one which makes\\ sense\end{tabular}                                                  & \begin{tabular}[c]{@{}l@{}}Statement1: He put a turkey into the fridge.\\ Statement2: He put an elephant into the fridge.\end{tabular}                                                                                                           \\ \hline
			Subtask B & \begin{tabular}[c]{@{}l@{}}Select the most corresponding\\ reason why this statement is\\ against common sense\end{tabular} & \begin{tabular}[c]{@{}l@{}}Statement: He put an elephant into the fridge.\\ A: An elephant is much bigger than a fridge.\\ B: Elephants are usually white while fridges \\ are usually white.\\ C: An elephant cannot eat a fridge.\end{tabular} \\ \hline
			Subtask C & \begin{tabular}[c]{@{}l@{}}Generate the reason why this\\ statement is against common\\ sense\end{tabular}                  & Statement: He put an elephant into the fridge.                                                                                                                                                                                                   \\ \hline
		\end{tabular}
		\caption{A brief summarization of Subtask A, B, and C in SemEval 2020 Task 4.}
		\label{tab:my-table}
	\end{table}
		%
	%
	
	\section{Related Work}
	
	Pre-trained Language models have been proved to be essential while dealing with sequence-to-sequence task. We present the language model we utilized in this section and demonstrate our consideration. We use RoBERTa \cite{DBLP:journals/corr/abs-1907-11692} for language comprehension task. The structure of RoBERTa is based on BERT \cite{devlin2018bert}. We also use BART \cite{lewis2019bart} for generation task. BART is a denoising autoencoder for pre-training sequence-to-sequence models. It uses the standard sequence-to-sequence Transformer architecture except the activation functions which is replaced by GeLUs. We find the structure of autoregressive decoder make the model can be directly ﬁne tuned for sequence generation tasks. BART is also enable to apply any type of document corruption which make the additional knowledge available to the model.
	
	Machine common sense has long been acknowledged as a critical component for natural language understanding. The challenge for the model has turned to abstractive knowledge comprehension. A task that is closely related to SemEval 2020 Task 4 is CommonsenseQA \cite{DBLP:journals/corr/abs-1811-00937}, in which the commonsense knowledge is required to make the correct prediction. In CommonsenseQA task, large-scale pre-trained models have brought significant performance gains. These gains are obtained by developing training strategies and enlarging training data \cite{DBLP:journals/corr/abs-1907-11692}, or improving parameter efficiency \cite{lan2019albert}. 
	
	While some of the improvement is achieved by developing the pre-trained model itself, some other approaches resort to external modules, e.g. knowledge extraction and graph-based reasoning in \cite{lv2019graphbased}. In our work, we are interested in the knowledge extraction method because developing the pre-trained model itself will be computationally expensive. 
	
	Moreover, different from recent common strategy \cite{lin2019kagnet,lv2019graphbased,ma2019generalizable} to use external knowledge which involves structured knowledge like ConceptNet \cite{DBLP:journals/corr/SpeerCH16}, we extract external knowledge from plain texts.

	\section{System Description}
	
	We first describe our evidence-searching approach which can be utilized in downstream tasks. Then we describe our systems for three subtasks.

	\subsection{Evidence-Searching Approach}
	\label{sec:evidence}
	
\begin{figure}
		\centering
		\includegraphics[width=0.90\textwidth]{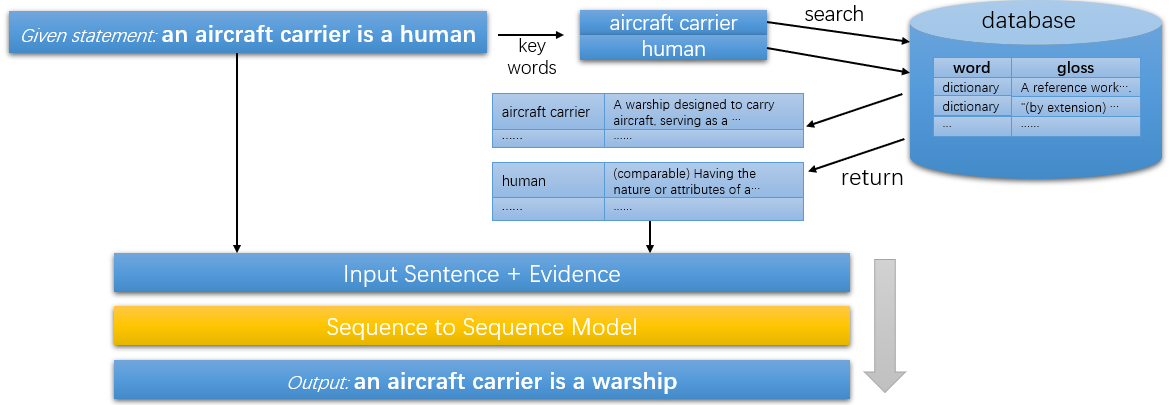}
		\caption{The Evidence-Searching Procedure}
		\label{image:evidence}
	\end{figure}

	As shown in Figure~\ref{image:evidence}, we first extract more than 1M two-element tuples (word, gloss) from Wiktionary\footnote{Wiktionary version: enwiktionary-20200220} with the help of Wiktextract package\footnote{https://github.com/tatuylonen/wiktextract} and adopt Elastic Search tools\footnote{https://www.elastic.co/} to index these tuples (note that the same word can have many meanings, thus in practice the tuples actually consist of 3 elements, adding an element which indicates the importance of this meaning). Then for each given statement, we extract its keywords with the help of Spacy\footnote{https://spacy.io/}. For each keyword, we search for those tuples whose "word" field matches the keyword. The Elastic Search engine ranks the matching score for tuples. We select top K tuples for each keyword. Thus the number of evidence tuples for a given statement is K*M (M denotes the number of keywords). In short, we search for the meaning of the keywords. Finally, the input sentences are produced by concatenating the original statement and corresponding evidence together. The detailed format is varied according to different subtasks and will be discussed later.

	\subsection{Systems for Subtask A\&B}
	\label{sec:subtaskAB}
	
	Both of subtask A\&B are to select one among several choices. Thus we implement two of the same model for subtaskA\&B. In a nutshell, our model is RoBERTa (A Robustly Optimized BERT Pretraining Approach)\cite{DBLP:journals/corr/abs-1907-11692} since it has been found that large-scale pre-trained contextualized representation masters a certain degree of commonsense knowledge \cite{zhou2019evaluating}. This is also supported by our experimental results.
	
	For subtask A, we adapt pre-trained $\mathsf{RoBERTa_{LARGE}}$ to subtask A dataset. We denote the hidden size for each layer (transformer blocks) as $H$, aggregate sequence representation as $C \in \mathbb{R}^{H}$ (final hidden state corresponding to the special [CLS] word embedding). The task-specific parameters we introduce is a vector $V \in \mathbb{R}^{H}  $. For each example, we simply input the two choice respectively and obtain the final aggregate representation $C_{i} \in \mathbb{R}^{H}$ for each choice $i$ whose dot product with the vector $V$ denotes a score for choice $i$. Thus the probability distribution is the softmax over the two choices:
	
	\begin{equation}
	\label{softmax}
	P_{i}=\frac{e^{V \cdot C_{i}}}{\sum_{j=1}^n e^{V \cdot C_{j}}}
	\end{equation}
	$n$ denotes the number of choices, which is 2 in subtask A. At testing time, the model's prediction $\hat{i}$ is the choice with the highest probability: 
	\begin{equation}
	\label{argmax}
	\hat{i}=\mathop{\arg\max}_{i} P_{i}
	\end{equation} 
	The model is trained with back propagation, using negative log-likelihood as loss function.
	
	In subtask B, given a statement that does not make sense, select the key reason from three options to explain why it does not make sense. Adapting RoBERTa to subtask B dataset is similar to the adaptation for subtask A. For each example, we construct three input sentences by concatenating statements with three choices respectively. Then input these sentences respectively. We compute the score for each of choice according to Equation~\ref{softmax} where n is 3 in subtask B. Then we follow the procedure described in subtask A.
	
	\subsection{System for Subtask C}
	\label{sec:subtaskC}
	
	Subtask C is an NLG (Natural Language Generation) task, which is quite different from subtask A\&B. Meanwhile, this subtask also requires commonsense knowledge and reasoning ability, which makes it more challenging. Our model for subtask C is BART (Denoising Sequence-to-Sequence Pre-training for Natural Language Generation, Translation, and Comprehension) \cite{lewis2019bart}. BART is a sequence-to-sequence model using a standard Transformer-based neural machine translation architecture. It is pre-trained by learning a model to reconstruct the original text from the corrupted text. It uses the same pre-training data as RoBERTa \cite{DBLP:journals/corr/abs-1907-11692}, consisting of 160Gb of news, books, stories, and web text.
	
	Specifically, we adapt pre-trained $\mathsf{BART_{LARGE}}$ to subtask C dataset. For each example, we follow our aforementioned evidence-searching approach to obtain evidence for the given statement. In the subtask C dataset, each statement has 3 referential reasons. During training, we construct 3 new example for each example in original dataset (e.g., for one original example $\mathsf{<statement, evidence, reason_1, reason_2, reason_3>}$ we construct three new example $\mathsf{<statement, evidence, reason_1>}$, $\mathsf{<statement, evidence, reason_2>}$, $\mathsf{<statement, evidence, reason_3>}$). Thus the total number of training examples is 3*N (N denotes the number of training examples originally). We denote this method as $\mathsf{Multi-target}$ training since the same input has multiple different targets. For each new example, the input sentence is the concatenation of the statement and evidence. Because BART has an autoregressive decoder, it can be directly fine-tuned for such a sequence generation task and can generate outputs autoregressively.

	\section{Experiments}
	
	We use the officially released dataset and standard train/trial/dev split of SemEval 2020 task 4 for experiments. We will give the performance of the best settings on test split. Note that we compare different settings through performance obtained by training the model on train\&trial split and testing it on dev split since testing on test split is inconvenient. We will also give our configuration of final submissions for subtask A, B, C in section~\ref{exp:subtaskA}, ~\ref{exp:subtaskB}, ~\ref{exp:subtaskC}
	
	\subsection{Experiments for Subtask A}
	\label{exp:subtaskA}
	
	We implement RoBERTa in FAIRSEQ \cite{ott-etal-2019-fairseq}. RoBERTa is optimized with Adam \cite{kingma2014adam} with the following parameters: $\beta_{1} = 0.9, \beta_{2} = 0.98, \epsilon = 1e-6$ and $L_{2}$ weight decay of 0.01. The learning rate is warmed up over the first 800 steps to a peak value of 1e-5, and then polynomially decayed. The clipping threshold of gradients is 0.1. RoBERTa is fine-tuned with a dropout of 0.1 on all layers and attention weights. It is fine-tuned for S=8,000 updates, with mini-batches containing B=8 sequences of maximum length T=512 tokens. 
	
	In experiments where we add evidence, for each statement we have several tuples (word, gloss). The evidence for the statement is in such format: "$\mathsf{<word_1>}$: $\mathsf{<gloss_1>}$ $\setminus$  $\mathsf{<word_2>}$: $\mathsf{<gloss_2>}$ $\setminus \cdots$". We construct two input sentences for each example in such format:  \textit{"$\mathsf{<Statement_1>}$ Context: $\mathsf{<Evidence_1>}$"}, \textit{"$\mathsf{<Statement_2>}$ Context: $\mathsf{<Evidence_2>}$"}. Note that due to the unavoidable memory limitation problem, we use memory efficient floating point numbers option provided by FAIRSEQ when we add evidence.
	
	We train the model on 1 $\times$ 11GB GeForce RTX 2080 GPU for around 15 minutes when we input statements without evidence and 2 $\times$ GPUs for around 100 minutes when we add evidence.
	
	We notice that there are some statements whose letters are all capitalized (e.g., A GIRL WON THE RACE WITH HER FRIEND) in Subtask A dataset. We capitalize the first letter and make other letters in lowercase. We denote this operation as $\mathsf{Lowercase}$ in Table~\ref{tab:subtaskA}.
	
	As shown in Table~\ref{tab:subtaskA}, we can see that the performance is slightly improved after we make some letter lowercase, since otherwise different forms of a word can be mapped to different embeddings while they have the same meaning. Then we explore the effect of evidence. By adding evidence to the input statement, we obtain a slightly better result in the development dataset while performance degradation is found in the test dataset. We hypothesize this discrepancy is because the amount of noise data in evidence is unstable.
	
	Our configuration of the final submission of subtask A has the same setting discussed in the first paragraph, with the lowercase operation, and without additional evidence. Our approach achieves the accuracy of 95.3\% on subtask A test dataset
	
	\begin{table}[]
		
		\begin{center}
			\begin{tabular}{lcc}
				\hline
				& Dev Acc & Test Acc \\
				$\mathsf{RoBERTa_{LARGE}}$ (our implementation) & 95.4    & -        \\
				+ Lowercase                                          & 95.7    & 95.3     \\
				+ Lowercase + Evidence                            & 96.3    & 94.1     \\ \hline
			\end{tabular}
			\caption{Model performance on the official subtask A development set and test set. Lowercase: make all the letters in lowercase except for the first letter. Evidence: add evidence from wiktionary}
			\label{tab:subtaskA}
		\end{center}
	\end{table}
	
	\subsection{Experiments for Subtask B}
	\label{exp:subtaskB}
	
	We primarily follow the optimization hyperparameters, given in Section~\ref{exp:subtaskA}, except for the batch size, number of warmup steps, and number of total updates which are 4, 500, and 10,000 separately. In the following experiments, we also follow the lowercase operation in Section~\ref{exp:subtaskA} as part of the default setting.
	
	In subtask A, given two similar statements, one makes sense while another one does not. In subtask B, the one that does not make sense is given, thus the other one---the statement that makes sense---can be used as a kind of evidence since the different words between two statements may be the keywords for explaining why given statement does not make sense. We denote this evidence as $\mathsf{Reasonable\;Statement}$, the evidence from wiktionary as $\mathsf{Wiktionary}$.
	
	In experiments where we do not add any evidence, we construct input sentences in such format \textit{"The statement '$\mathsf{<Statement>}$' is absurd. Because $\mathsf{<Choice_{i}>}$"} in which we concatenate some additional words (\textit{"The statement '$\cdots$' is absurd. Because $\cdots$"}) to the sentence and denote this technique as $\mathsf{Extra\;Words}$. Moreover, in experiments where we add the reasonable statement, the input format for each choice is \textit{"Reasonable statement: $\mathsf{<Reasonable\;Statement>}$ $\setminus$ The statement '$\mathsf{<Statement>}$' is absurd. Because $\mathsf{<Choice_{i}>}$"}. If the evidence is added, then the format will be \textit{"Context: $\mathsf{<Wiktionary>}$ Reasonable statement: $\mathsf{<Reasonable\;Statement>}$ $\setminus$ The statement '$\mathsf{<Statement>}$' is absurd. Because $\mathsf{<Choice_{i}>}$"}
	
	In Table~\ref{tab:subtaskB} we present the result of different settings. We see that the extra words can bring 1.6\% absolute improvement since they indicate that the given statement does not make sense and the choice is the reason for that. We also see that using corresponding reasonable statement and wiktionary evidence achieve comparable performance while they involve extra computational cost. Therefore, we only use extra words in the final submission of subtask B and achieve accuracy of 93.2\%
	
	\begin{table}[]
		
		\begin{center}
			\begin{tabular}{lcc}
				\hline
				& Dev Acc & Test Acc \\
				$\mathsf{RoBERTa_{LARGE}}$ (our implementation) & 91.6    & -        \\
				+ Extra Words                                      & 93.2    & 93.2     \\
				+ Extra Words + Reasonable Statement             & 93.2    & -     \\
				+ Extra Words + Wiktionary                            & 92.4    & -     \\ 
				+ Extra Words + Reasonable Statement + Wiktionary     & 93.1    & -     \\
				\hline
			\end{tabular}
			\caption{Model performance on the official subtask B development set and test set. Extra Words: extra words indicating given statement does not make sense. Reasonable Statement: the corresponding reasonable statement. Wiktionary: the obtained evidence from wiktionary}
			\label{tab:subtaskB}
		\end{center}
	\end{table}
	
	\subsection{Experiments for Subtask C}
	\label{exp:subtaskC}
	
	BART is also implemented in FAIRSEQ \cite{ott-etal-2019-fairseq}. It is optimized with Adam \cite{kingma2014adam} with the following parameters: $\beta_{1} = 0.9, \beta_{2} = 0.999, \epsilon = 1e-8$ and $L_{2}$ weight decay of 0.01. The learning rate is warmed up over the first 500 steps to a peak value of 3e-5, and then polynomially decayed. The clipping threshold of gradients is 0.1. BART is fine-tuned with a dropout of 0.1 on all layers and attention weights. It is fine-tuned for S=1,200 updates, with mini-batches containing B=32 sequences of maximum length T=512 tokens. Note that we keep the same experiment settings in the following experiments. We train the model on 4 $\times$ 11GB GeForce RTX 2080 GPU, for 15 minutes to 1 hour according to different settings.
	
	In the following experiments, we follow the input format described in Section~\ref{exp:subtaskB}, removing the "$\mathsf{<Choice_{i}>}$" only. We conduct experiments on different combinations of the methods described above (Multi-target: Section~\ref{sec:subtaskC}; Extra Words, Reasonable Statement, and Wiktionary: Section~\ref{exp:subtaskB}) and explore the effect of them.
	
	As shown in Table~\ref{tab:subtaskC}, by using Multi-target training, we can obtain a 4.51 improvement on BLEU score. Compare to the baseline where we simply use the first referential answer of each example in training data as the target of the model output, the Multi-target method provides a larger amount of training data and thus helps the model get better performance. From Table~\ref{tab:subtaskC} we can see performance degradation appears as we add some extra material but then performance improved as we add more material. We hypothesize the degradation is because the complexity of the input sentence increases as we add extra material. When Extra Words, Reasonable Statement, and Wiktionary are all added, the benefits outweigh the disadvantages they bring. Therefore, we use all of them in the final submission of subtask C and achieve the BLEU score of 18.5 (3rd place) and human evaluation score of 2.08 (2nd place), which obtains a 0.14 gain over 3rd place and only 0.02 less than 1st place.
	
	The best score (20.39) on the test set in Table~\ref{tab:subtaskC} outperforms the score we achieved during the competition (18.5). Note that we might achieve a better human evaluation score accordingly. There are two reasons for that. Firstly, we optimize our evidence-searching approach after the competition, improving the quality of the evidence (All the experiments with evidence added in this paper are using the improved version. Details of optimizing method are shown in Appendices). Secondly, we observe that during the training process, the model performs well at the beginning but turns to mess later. Thus it's difficult to choose the best model during the training process when we cannot evaluate it on the test set. When the competition has ended, however, we can evaluate our models and choose the best one during the training process. 
	
	\begin{table}[]
		
		\begin{center}
			\begin{tabular}{lcc}
				\hline
				& Dev BLEU & Test BLEU \\
				$\mathsf{BART_{LARGE}}$ (our implementation)     & 15.15    & -        \\
				+Multi-target                                      & 19.66    & 19.43     \\
				+Multi-target + Extra Words                       & 18.10    & -     \\
				+Multi-target + Wiktionary                       & 18.98    & -         \\
				+Multi-target + Extra Words + Wiktionary         & 19.50    & -      \\ 
				+Multi-target + Extra Words + Reasonable Statement         & 18.98    & -      \\
				+Multi-target + Extra Words + Reasonable Statement + Wiktionary & 20.03  & 20.39  \\
				\hline
			\end{tabular}
			\caption{Model performance on the official subtask C development set and test set. Multi-target: the same input has multiple different targets. Extra Words: extra words indicating given statement does not make sense. Reasonable Statement: the corresponding reasonable statement. Wiktionary: the obtained evidence from wiktionary}
			\label{tab:subtaskC}
		\end{center}
	\end{table}
	
	\section{Conclusion}
	\label{sec:length}
	
	In this work, we choose the different large-scale pre-trained models as the backbone for three subtasks and propose a novel way to search for evidence, which aims to obtain the meaning of the keywords in the given statement. Our experiments demonstrate the importance of additional knowledge for language models to understand the content. The results show that our evidence-searching approach is helpful to commonsense explanation task.

	\bibliographystyle{coling}
	\bibliography{semeval2020}

	\section*{Appendices}
	We optimize our evidence-searching approach after the competition. That is, filtering out some meaningless word-gloss pairs while keeping the framework unchanged. We observe two kinds of meaningless pairs in the evidence. First, pairs contain misleading gloss, e.g., when we search the gloss for the keyword "car", we obtain "CAR: initialism of 'Central African Republic' \ CAR: (uncountable chemistry) abbreviation of 'carnitine'"; when we search the gloss for "like", "like like: (slang) To fancy; to be attracted to" is obtained. To address this problem, we remove the pairs whose gloss contains "initialism/historical/obsolete/abbreviation/(dated)/slang/acronym/(US)/synonym/archaic/surname/(rare)", and remove the pairs whose word contains "-" or capital letters. Replacing these with other pairs. Second, we also observe many pairs only contain prototype information. For example: "watermelons: plural of 'watermelon'", "concentrated: past of 'concentrate'", "facebook: alternative form of 'Facebook'". To obtain more meaningful gloss, we search the pairs for the prototype word, regarding the prototype word as one of the keywords. Specifically, we detect whether the gloss contain words like "plural of/past of/third person singular of/clipping of/alternative form of/alternative spelling of" which indicate the gloss points to a prototype word. In that case, we use the same search function to acquire evidence for prototype word and incorporate the new evidence. (note that to avoid infinite loop, we do not detect the prototype in sub-search). Besides, we also adjust the number of evidence tuples for a keyword dynamically. For the statement contains less keywords, we obtain more evidence tuples for each keywords. Thus the length of the evidence for statements will be more stable.

\end{document}